\documentclass[sigconf]{acmart}
\usepackage{balance}

\AtBeginDocument{%
  \providecommand\BibTeX{{%
    \normalfont B\kern-0.5em{\scshape i\kern-0.25em b}\kern-0.8em\TeX}}}

\setcopyright{acmcopyright}
\copyrightyear{2022}
\acmYear{2022}
\acmDOI{10.1145/3511808.3557152}

\acmConference[CIKM ’22]{Proceedings of the 31st ACM International Conference on Information and Knowledge Management}{October 17--21, 2022}{Atlanta, GA, USA.}

\acmBooktitle{Proceedings of the 31st ACM International Conference on Information and Knowledge Management (CIKM '22), October 17--21, 2022, Atlanta, GA, USA}

\acmPrice{15.00}
\acmISBN{978-1-4503-9236-5/22/10}

\begin{document}

\title{Billion-user Customer Lifetime Value Prediction: An Industrial-scale Solution from Kuaishou}


\author{Kunpeng Li}
\orcid{0000-0002-8713-281X}
\affiliation{
  \institution{Beijing Kuaishou Technology Co., Ltd.}
  \city{Beijing}
  \country{China}
}
\email{likunpeng@kuaishou.com}

\author{Guangcui Shao}
\orcid{0000-0002-3098-3621}
\affiliation{
  \institution{Beijing Kuaishou Technology Co., Ltd.}
  \city{Beijing}
  \country{China}
}
\email{shaoguangcui@kuaishou.com}

\author{Naijun Yang}
\orcid{0000-0003-1837-8477}
\affiliation{
  \institution{Beijing Kuaishou Technology Co., Ltd.}
  \city{Beijing}
  \country{China}
}
\email{yangnaijun@kuaishou.com}

\author{Xiao Fang}
\orcid{0000-0002-0652-0977}
\affiliation{
  \institution{Beijing Kuaishou Technology Co., Ltd.}
  \city{Beijing}
  \country{China}
}
\email{fangxiao05@kuaishou.com}

\author{Yang Song}
\orcid{0000-0002-1714-5527}
\authornote{Corresponding author.}
\affiliation{
  \institution{Beijing Kuaishou Technology Co., Ltd.}
  \city{Beijing}
  \country{China}
}
\email{yangsong@kuaishou.com}

%

\renewcommand{\shortauthors}{Kunpeng Li et al.}

\begin{abstract}
Customer Life Time Value (LTV) is the expected total revenue that a single user can bring to a business. It is widely used in a variety of business scenarios to make operational decisions when acquiring new customers. Modeling LTV is a challenging problem, due to its complex and mutable data distribution. Existing approaches either directly learn from posterior feature distributions or leverage statistical models that make strong assumption on prior distributions, both of which fail to capture those mutable distributions. In this paper, we propose a complete set of industrial-level LTV modeling solutions. Specifically, we introduce an \textbf{O}rder \textbf{D}ependency \textbf{M}onotonic \textbf{N}etwork (ODMN)  that models the ordered dependencies between LTVs of different time spans, which greatly improves model performance. We further introduce a \textbf{M}ulti \textbf{D}istribution \textbf{M}ulti \textbf{E}xperts (MDME) module based on the \textit{Divide-and-Conquer} idea, which transforms the severely imbalanced distribution modeling problem into a series of relatively balanced sub-distribution modeling problems hence greatly reduces the modeling complexity. In addition, a novel evaluation metric \textit{Mutual Gini} is introduced to better measure the distribution difference between the estimated value and the ground-truth label based on the Lorenz Curve. The ODMN framework has been successfully deployed in many business scenarios of Kuaishou, and achieved great performance. Extensive experiments on real-world industrial data demonstrate the superiority of the proposed methods compared to state-of-the-art baselines including ZILN and Two-Stage XGBoost models.
\end{abstract}


\maketitle

\section{Introduction}
Customer Life Time Value (LTV) refers to the sum of all the economic benefits a company gets from all the interactions during the user's lifetime. With the development of the modern economy, companies increasingly benefit from building and maintaining long-term relationships with customers. Under this circumstance, it is particularly important to make operational decisions based on customer lifetime value. For instance, marketers need to accurately predict the total consumption income of customers for a long time in the future, ranging from a few months to multiple years, so as to make a reasonable financial budget planning and guidelines to carry out the customer relationship management (CRM). In addition, for many companies' growth operations and advertising businesses, it is necessary to predict the long-term contribution of the users to the platform for calculating the ROI of the investment, in order to guide the selection of the best channel for delivery, or to bid according to the quality.

In the past few years, a large number of academic literature on modeling LTV has been published, most of which can be divided into two categories. The first category of methods leverages historical experience or classical probability and statistical models \cite{gupta2006modeling, chang2012customer, talaba2013comparison}. The other is to model LTV directly through machine learning or deep learning \cite{drachen2018or, vanderveld2016engagement, chen2018customer}.

\begin{figure}[t]
  \centering
  \includegraphics[width=3.0in]{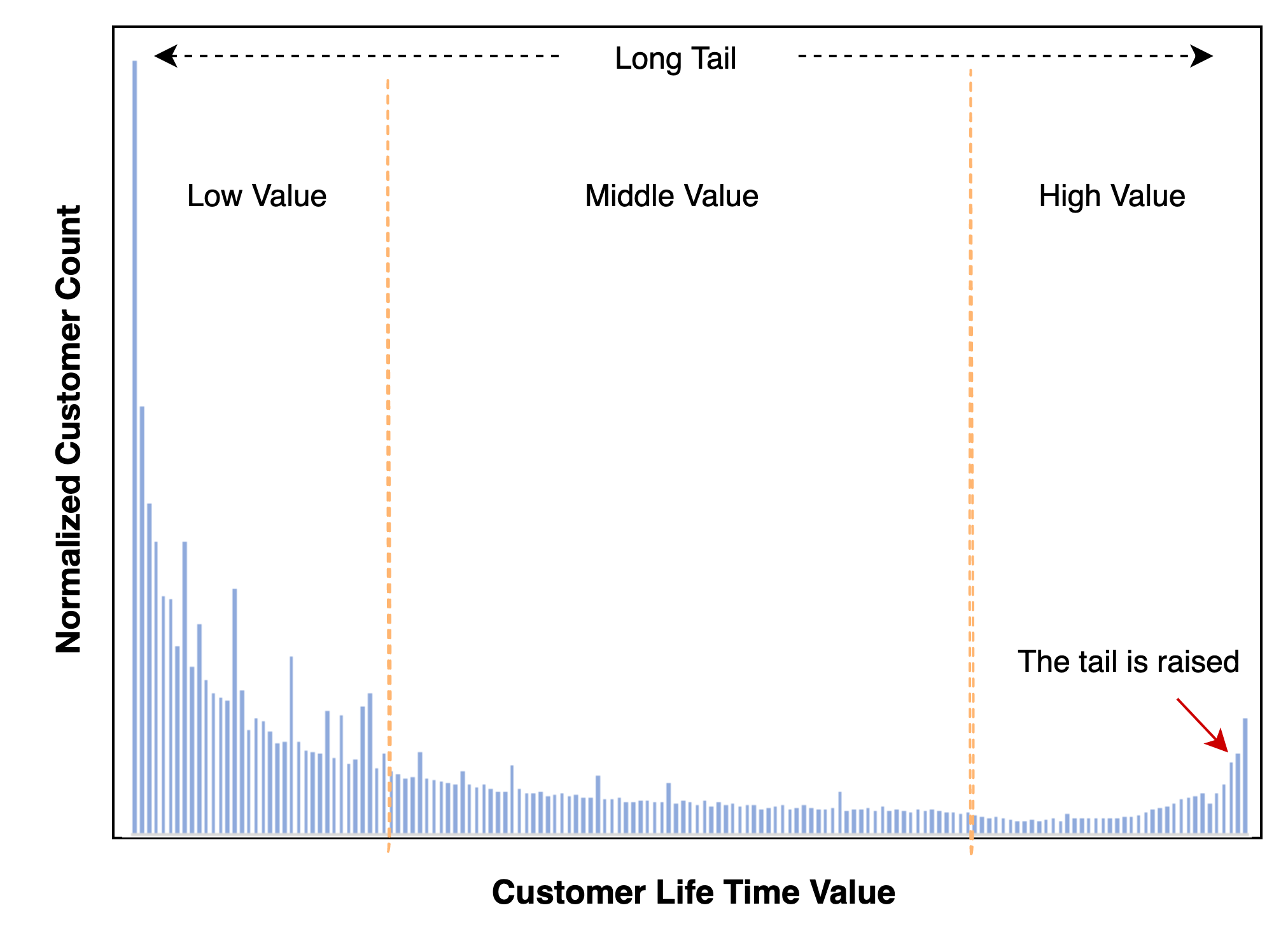}
  \caption{The distribution of LTV in our dataset, illustrating the sparsity of high-value users and the atypical long-tail issues. In particular, the distribution has a raised tail.}
  \Description{data}
  \label{fig:data}
\end{figure}

Although existing methods greatly improve the performance of LTV predictions, there are still two major challenges to be solved. Firstly, they do not consider how to deal with the complex distribution of LTV more effectively for obtaining better benefits. Conventional methods are based on Mean Square Error(MSE) to measure loss, but the square term is very sensitive to large values. Under the training algorithm based on stochastic gradient descent, it is easy to generate a large gradient vector, which affects the convergence and stability of the model. Figure \ref{fig:data} shows the LTV distribution of Kuaishou growth business, where the LTV is defined as the total revenue contributed by per user to Kuaishou through e-commerce, advertising and other channels in the future. It is an atypical long-tailed distribution, which is hard to fit well using the aforementioned methods. Secondly, they also ignore the ordered dependencies between LTVs of different time spans. In many business scenarios, in addition to modeling long-term user value, we also pay attention to short-term user value to facilitate timely adjustment of operation decisions, that is, we need to model multi-time-span user values. For example, $ltv_{30}$, $ltv_{90}$, $ltv_{180}$ and $ltv_{365}$ represent the value that users will contribute to the platform in the next month, quarter, half a year and year respectively. There are two points to note: (1). From a business perspective, LTVs with different time spans are subject to ordered dependencies, that is, $ltv_{30} \le ltv_{90} \le ltv_{180} \le ltv_{365}$. (2). From the perspective of modeling difficulty, the longer the time span, the greater the modeling difficulty of LTV. It is obvious that $ltv_{30}$ is expected to be predicted more accurately than $ltv_{365}$. Conventional solutions either model LTV with different time spans separately, which is costly, or through multi-task learning. However, they ignore the ordered dependencies between LTV tasks with different time spans. A typical bad case is that the $ltv_{30}$ estimated by the model may be greater than $ltv_{90}$, which is not in line with business expectations. In order to alleviate this problem, we propose the \textbf{O}rder \textbf{D}ependency \textbf{M}onotonic \textbf{N}etwork (ODMN) framework, which explicitly models the ordered dependencies between LTV tasks with different time spans, and assists the learning of more difficult and complex long-term LTV tasks through easier-to-model short-term LTV tasks, thereby achieving a great improvement in model performance. In addition, for each specific LTV prediction task, we propose a \textbf{M}ulti \textbf{D}istribution \textbf{M}ulti \textbf{E}xperts (MDME) module based on the \textit{Divide-and-Conquer} idea. Through the distribution segmentation and sub-distribution bucketing mechanism, the seriously imbalanced distribution modeling problem is transformed into a series of more balanced sub-distribution modeling problems, which greatly mitigates the modeling difficulty and enhances the performance of LTV prediction.

This paper continues to explore along the line of modeling LTV with machine learning methods, in particular deep learning models. To summarize, the main contributions are fourfold:

\begin{itemize}
  \item We propose a novel and effective architecture named Order Dependency Monotonic Network (ODMN), which actively captures the ordered dependencies between LTVs of different time spans to enhance prediction performance.
  \item A module named Multi Distribution Multi Experts (MDME) based on the idea of \textit{Divide-and-Conquer} is designed to deal with the complex and imbalanced distribution of LTV from coarse-grained to fine-grained, reducing modeling difficulty and improving performance.
  \item The proposal of \textit{Mutual Gini} on the basis of the Lorenz Curve quantitatively measures the model's ability to fit an imbalanced label distribution for the first time.
  \item The method in the paper has strong scalability and has been deployed to the company's online business. Both offline experiments and online A/B tests have shown the effectiveness of our approach.
\end{itemize}

\section{Related Work}


\subsection{LTV Prediction}
The existing LTV prediction methods are mainly divided into two lines. The first is based on historical data or classical probability and statistics model. The RFM framework \cite{fader2005rfm}, groups users based on recency, frequency, and monetary value of historical consumption, to roughly calculate how long or how often the users buy or consume something. The basic assumption it follows is that the users who tend to consume more frequently, will be more likely to consume again in the near future if they have relatively higher consumption recently. The BTYD \cite{wadsworth2012buy} family is a very classic probability model for user repeat purchase/churn, which assumes that both users churn and purchase behavior follow some sort of stochastic process, one of the most well-known solutions is the Pareto/NBD \cite{fader2005rfm} model, which is commonly used in non-contractual, continuous consumption scenarios (that is to say, customers may purchase at any time). This method will model two-parameter distributions, Pareto distribution is used for binary classification, predicting whether users are still active, and negative binomial distribution is used to estimate the frequency of purchases. Most of these methods are coarse-grained modeling of the consumption habits of users, such as whether to purchase it, the frequency of purchases, etc, they do not provide fine-grained modeling of the amount spent or the specific value that users will contribute to the platform over a long period of time. The second is to model LTV directly through machine learning or deep learning. Early methods rely more on hand-crafted features and tree-structured models. Recently, deep learning technology has also been applied to LTV prediction. \cite{drachen2018or} is a two-stage XGBoost model, which first identifies high-quality users and then predicts their LTV. \cite{vanderveld2016engagement} firstly splits users into several groups and performs random forest prediction within each group. ZILN \cite{wang2019deep} assumes that it obeys the log-normal distribution, and exploits the advantages of DNN in feature intersection and information capture to fit the mean and standard deviation of the distribution, and finally the expected value of the log-normal distribution is used as an estimate of LTV, but the distribution assumption is so restrictive, which leads to limitations in its application. \cite{chen2018customer} shows that CNN is more efficient than MLP to model time series data for LTV prediction. \cite{xing2021learning} focuses on how to use the wavelet transform and GAT to learn more reliable user representation from the sparse and changeable historical behavior sequence of users. However, they are still based on simple MSE loss to fit LTV distribution.

\subsection{Long Tail Modeling}
In the real world, most users are low-value contributors, so the distribution of LTV data is extremely imbalanced, of which long-tailed distributions are the most common. As we all know, modeling long-tailed distributions is notoriously difficult to solve. Research in this area is roughly divided into three categories: Class Re-Balancing, Information Augmentation and Module Improvement \cite{zhang2021deep}.

\begin{figure*}[h]
  \centering
  \includegraphics[width=\linewidth]{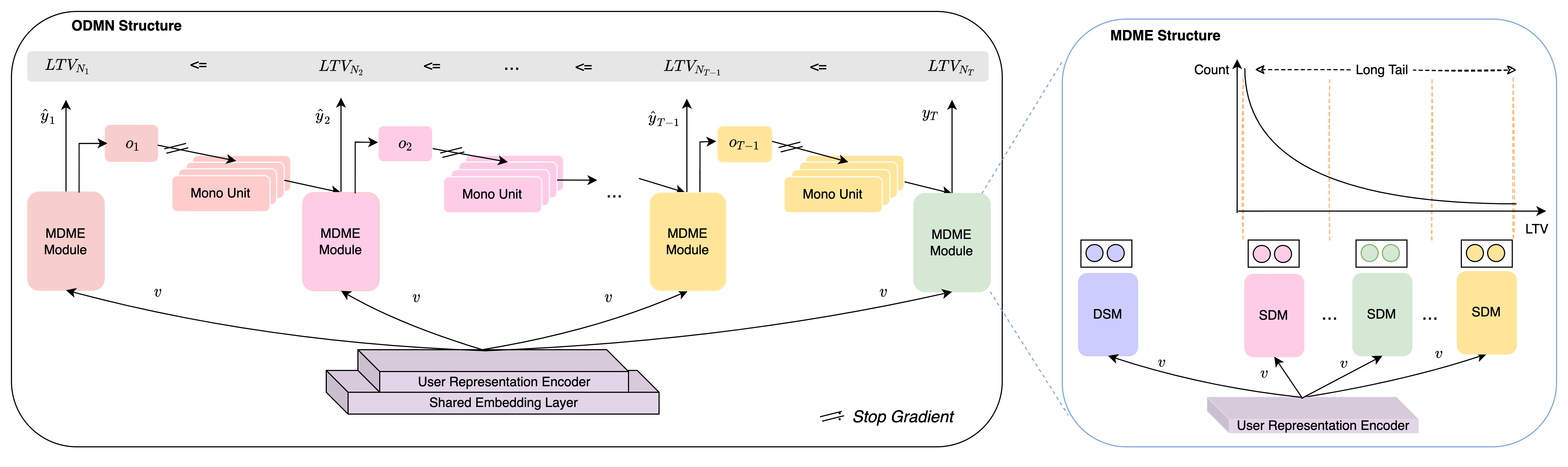}
  \caption{The overall structure of our proposed ODMN. The LTV of each time span is modeled by a MDME module, $o_t$ represents the normalized bucket multinomial distribution. The \textit{Mono} \textit{Unit} is a MLP with non-negative parameters, which can capture the shift trend of the normalized multinomial distribution of the upstream tasks. Then the downstream tasks can perceive the distribution changes of the upstream tasks, so as to make corresponding distribution adjustments.}
  \Description{The structure of ODMN}
  \label{fig:ODMN}
\end{figure*}

\begin{figure*}[h]
  \centering
  \includegraphics[width=\textwidth]{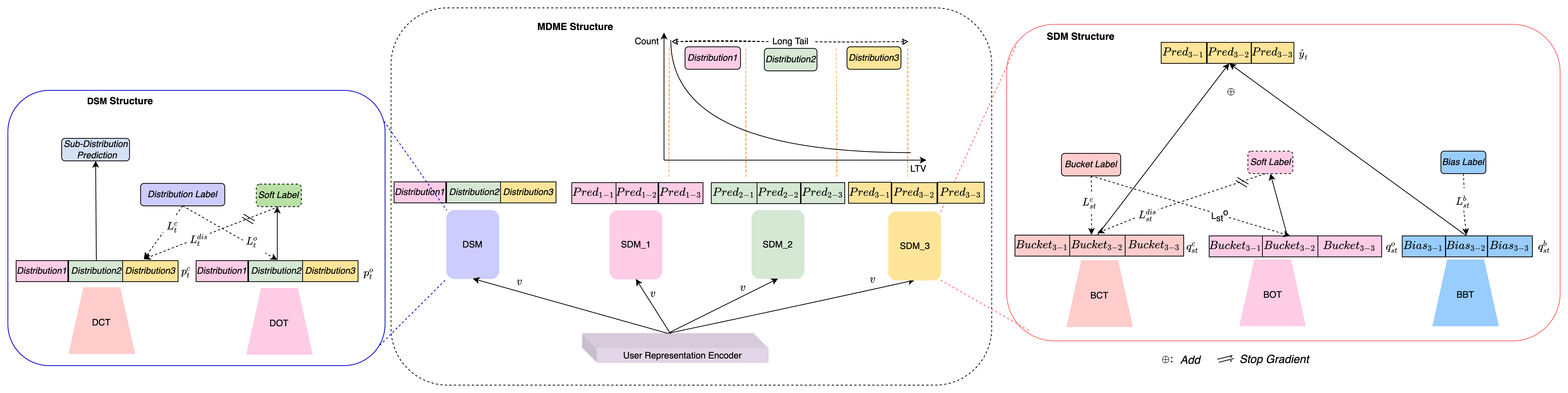}
  \caption{The structure details in MDME module. DSM denotes distribution segmentation module, to segment the entire LTV distribution. SDM denotes sub-distribution modeling module. DCT and DOT in DSM denote distribution classification tower and distribution ordinal tower, respectively. BCT, BOT and BBT in SDM denote bucket classification tower, bucket ordinal tower and bucket bias tower, respectively. See Section \ref{section:MDME}  for details.}
  \Description{The structure of MDME}
  \label{fig:MDME}
\end{figure*}

In Class Re-Balancing, one of the most widely used methods to solve the problem of sample imbalance is Re-Sampling, including Random Over-Sampling (ROS) and Random Under-Sampling (RUS). ROS randomly repeats the samples of the tail class, while RUS randomly discards the samples of the head class. When the class is extremely imbalanced, ROS tends to overfit the tail class, while RUS will reduce the model performance of the head class. Another conventional approach to this problem is Cost-Sensitive Learning, also known as Re-Weight. This method assigns larger weights to minor samples and smaller weights to major samples. However, the weights often depend on the prior distribution of the sample label, and when the distribution changes, the sample weights need to be re-determined. 

Information Augmentation mainly alleviates the long-tailed problem through transfer learning and data augmentation. For transfer learning, it includes methods such as head-to-tail knowledge transfer \cite{yin2019feature}, knowledge distillation \cite{he2021distilling} and model pre-training \cite{erhan2010does}, while data augmentation is to alleviate the problem of data imbalance by enhancing quantity and quality of sample sets, but the quality of expanded data is difficult to guarantee.

Module Improvement is to improve the modeling ability of long-tailed distribution from the perspective of network structure optimization. It can be roughly divided into the following categories: (1). Improving feature extraction through representation learning. (2). Designing classifiers to enhance the classification ability of the model. (3). Decoupling feature extraction and classifier classification. (4). Ensemble learning to improve the entire architecture.

Our work continues the line of modeling LTV through machine learning/deep learning. We effectively model LTV from the perspective of imbalanced distribution processing, and model the ordered dependencies of multi-time-span LTVs for the first time, making important technical contributions to the field of LTV modeling.

\section{Method}
In this section, we will present a novel LTV modeling approach, \textbf{O}rder \textbf{D}ependency \textbf{M}onotonic \textbf{N}etwork (ODMN), which simultaneously models multiple LTV objectives with spans in time. As shown in Figure \ref{fig:ODMN}, it is a classic share-bottom multi-task learning framework, which can be upgraded to a more complex structure, such as Multi-gate Mixture-of-Experts (MMoE) \cite{ma2018modeling} or Progressive Layered Extraction (PLE) \cite{tang2020progressive}, according to the business situations and requirements. These backbone networks are beyond the focus of this paper.

\subsection{Problem Formulation}
Given the feature vector $\textbf{x}$ under a fixed time window of the user (e.g. 7 days), predicting the value it will bring to the platform in the next $N$ days ($LTV_{N}$). Obviously, $LTV_{N-\Delta}$ $\leq$ $\cdots$ $\leq$ $LTV_N$ $\leq$ $\cdots$ $\leq$ $LTV_{N+\Delta}$. The multi-task framework needs to predict LTV of $T$ different time spans based on the input feature $\textbf{x}$ at the same time:
\begin{equation}
\begin{split}
  \hat{y}=f \left( LTV_{N_1}, LTV_{N_2}, \cdots ,  LTV_{N_t}, \cdots ,  LTV_{N_T}| \textbf{x} \right) \\
  1 \leq t \leq T, N_1 < N_2 < \cdots < N_t < \cdots < N_T.
\end{split}
\end{equation}

\subsection{Shared Embedding Layer and User Representation Encoder}
The features of model input mainly include user portrait characteristics, historical behavior information during a period of time window (e.g. 7 days), etc. For the historical behavior information of users, we organize them into day-level sequence format, such as "active duration": [13, 7, 70, 23, 44, 12, 9], which represents the active minutes one user interacts with the Kuaishou APP per day in the past 7 days. From the practical effect, the behavior features of day-level sequence format can bring more information gain. At the same time, in order to improve the generalization ability of the model, we do not consider highly personalized features such as $user\_id$, $item\_id$ in feature selection. In offline data analysis, we found that the LTV of some business scenarios has typical cyclical fluctuation due to user's periodical behavior. For example, the consumption tendency of users on weekends increases, while decreases on weekdays. In addition, LTV is also more sensitive to seasonal or abnormal events. For instance, during holidays or e-commerce shopping festivals, the user's consumption tendency increases significantly. The model needs to capture these signals, so we collect these periodic, seasonal or abnormal event information as parts of the features input to the model. Whether in marketing, advertising or other businesses, we often pay attention to the LTV prediction performance of users from different channels, therefore, some channel-related information can be added to enhance the model's ability to predict LTV of channel dimension. Drawing on the idea of the RFM method, we also introduce some recency, frequency, and monetary value related to the user's historical consumption as features.

As shown in Figure \ref{fig:ODMN}, given the input features $\textbf{x}$, we embed each entry $x_i$ ( $x_i$ $\in$ $\textbf{x}$, 1 $\leq$ $|\textbf{x}|$) into a low dimension dense vector representation $v_i$ via a \textit{Shared} \textit{Embedding} \textit{Layer}. Among them, the features of the dense type will be discretized firstly, for some dense features of long-tailed distribution, we tend to process at equal frequencies. The embedding vector of each feature entry will be concatenated together as the final user representation $v$ = [$v_1$;$v_1$;$\cdots$;$v_{|\textbf{x}|}$], where [$\cdot$;$\cdot$] denotes the concatenation of two vectors. Better user representation can be obtained through more complex feature intersections (e.g. DeepFM \cite{guo2017deepfm}, DCN \cite{wang2017deep}) or user behavior sequence modeling (e.g. Transformer \cite{vaswani2017attention}, Bert \cite{devlin2018bert}), but it is not the focus of this paper.

\subsection{Multi Distribution Multi Experts Module}
\label{section:MDME}
For each specific LTV prediction task, we design a novel module MDME (\textbf{M}ulti \textbf{D}istribution \textbf{M}ulti \textbf{E}xperts), of which the core idea is \textit{Divide-and-Conquer}. By decomposing the complex problem into several simpler sub-problems, each of them can be broken down easily. It is true that modeling LTV directly will bring a lot of troubles, among which the most important one is that, due to the imbalance distribution of samples, it is brutal to learn the tailed samples well. The sample distribution mentioned here refers to the distribution of LTV value. In theory, it is less challenging for the model to learn from a more balanced sample distribution than from an imbalanced one \cite{xiang2020learning}. Inspired by this, we try to "cut" the entire sample set into segments according to the LTV distribution, so that the imbalanced degree of LTV distribution in each segment is greatly alleviated.

Figure \ref{fig:MDME} shows the structure of MDME. The whole end-to-end learning process consists of three stages. Firstly, the \textbf{D}istribution \textbf{S}egmentation \textbf{M}odule (DSM) realizes the segmentation of LTV distribution, that is, sub-distribution multi-classification. Specifically, the sample set is divided into several sub-distributions according to the LTV value. For some LTV values that account for a very high proportion in the sample set, they can be regarded as an independent sub-distribution, just like 0 in the zero-inflated long-tailed distribution. After distribution segmentation, each sample will only belong to one of the sub-distributions. We need to learn the mapping relationship between the samples and the sub-distributions, which is a typical multi-classification problem. At this stage, the sample label is defined as the sub-distribution number. Secondly, we continue to adhere to the basic principle of \textit{Divide-and-Conquer}, and use the \textbf{S}ub-\textbf{D}istribution modeling \textbf{M}odule (SDM) to further break each sub-distribution into several buckets according to the actual LTV value of the samples within the sub-distribution, then we transform the sub-distribution modeling into the problem of multi-bucket classification. In this way, we can adjust the number of samples falling into each bucket by tuning the bucket width, thus, the number of samples in each bucket is approximately equal. The sample label at this stage is the bucket number. Similar to distribution segmentation, the LTV value with a high proportion of samples in the sub-distribution can be used as an independent bucket number, that is, the corresponding bucket width is 1. After two stages of label redefinition, the modeling difficulty of the entire LTV distribution is greatly mitigated and the modeling granularity has been reduced to each bucket. At this time, the LTV distribution of samples in each bucket has been relatively balanced. The last stage is to learn a bias in the bucket to achieve fine-grained LTV modeling in the bucket, which is also processed by SDM. We perform min-max normalization on the LTV value of the samples in the bucket, so that the range of LTV value is compressed to between 0 and 1, which we define as the \emph{bias} \emph{coefficient}. The bias coefficient is then regressed based on MSE. Since MSE is very sensitive to extreme values, normalization can restrict the value range of label and the magnitude of loss, so as to reduce the interference to the learning of other tasks. In summary, for each sample, the target sub-distribution is first determined by DSM, then SDM corresponding to the target sub-distribution estimates the target bucket and the bias coefficient in the bucket to obtain the final LTV. The whole process realizes the prediction of LTV from coarse-grained to fine-grained.

DSM includes the \textbf{D}istribution \textbf{C}lassification \textbf{T}ower (DCT) and the \textbf{D}istribution \textbf{O}rdinal \textbf{T}ower (DOT), SDM consists of \textbf{B}ucket \textbf{C}lassification \textbf{T}ower (BCT), \textbf{B}ucket \textbf{O}rdinal \textbf{T}ower (BOT) and \textbf{B}ucket \textbf{B}ias \textbf{T}ower (BBT), of which DCT and BCT have similar structures, they respectively implement multi-classification of sub-distribution and bucket multi-classification within sub-distribution. The activation function of the output layer in both DCT and BCT is \emph{softmax}, which generates sub-distribution multinomial distribution and bucket multinomial distribution. We use the estimated probability of each sub-distribution in DCT as the weight of the bucket multinomial distribution within each sub-distribution to obtain a normalized bucket multinomial distribution of the entire LTV, denoted as $o_{t}$, where $t$ represents the $t_{th}$ LTV prediction task. The ODMN framework performs linear transformation on the normalized bucket multinomial distribution of the upstream output through Mono Unit, and directly adds the transformation result to the output logits of downstream DCT and BCT, which can affect the output distribution of the downstream task, so that once the normalized bucket multinomial distribution of the upstream output shifts, the downstream task can capture this signal in time, and the output distribution will also shift in the same direction accordingly. The normalized bucket multinomial distribution and Mono Unit will be introduced in next subsection. In addition, there is obviously a relative order relationship between sub-distributions and between buckets. We found that introducing a module to explicitly learn this relationship can further improve the accuracy of classification and ranking. Therefore, the DOT and the BOT are designed based on Ordinal Regression \cite{fu2018deep} to model this kind of order relationship, of which, the output layer activation function is set to \emph{sigmoid}, and the estimated values of DOT and BOT will be used as \emph{soft} \emph{label} to guide the learning of DCT and BCT through distillation respectively. After DCT and BCT determining the target sub-distribution and the corresponding target bucket, it's time for BBT to implement the fine-grained regression by fitting the normalized bias of LTV in each bucket based on MSE, the number of logits in the output layer of BBT is set to the number of buckets whose width is greater than 1, and \emph{sigmoid} is selected as the activation function of the output layer to limit the estimated value between zero and one. Assuming that the $t_{th}$ task divides LTV into $S_t$ sub-distributions, for each one $s_t$(1 $\leq$ $s_t$ $\leq$ $S_t$), the output function of each tower module is defined as: 

\begin{small}
\begin{equation}
\begin{split}
  p_t^c=softmax \left( f_{DCT}^t \left( v \right) + 
  Mono_{DCT} \left( stop\_gradient \left( o_{t-1} \right) \right) \right)
\end{split}
\end{equation}
\end{small}

\begin{small}
\begin{equation}
\begin{split}
   q_{s_t}^c=softmax \left( f_{BCT}^{s_t} \left( v \right) + 
   Mono_{BCT} \left( stop\_gradient \left( o_{t-1} \right) \right) \right)
\end{split}
\end{equation}
\end{small}

\begin{equation}
  p_t^o=sigmoid\left( g_{DOT}^t \left( v \right)  \right)
\end{equation}

\begin{equation}
  q_{s_t}^o=sigmoid \left( g_{BOT}^ {s_t} \left( v \right) \right)
\end{equation}

\begin{equation}
  q_{s_t}^b=sigmoid \left( h_{BBT}^ {s_t} \left( v \right) \right),
\end{equation}

\noindent
$f_{DCT}^t \left( \cdot \right)$ and $g_{DOT}^t \left( \cdot \right)$ are the functions of the DCT and the DOT modules for the $t_{th}$ task respectively, $p_t^c,p_t^o \in \mathbb{R} ^k$, and $k$ are the dimensions of the tower output, that is, the number of sub-distributions divided. $f_{BCT}^{s_t} \left( \cdot \right)$ , $g_{BOT}^{s_t} \left( \cdot \right)$ , $h_{BBT}^{s_t} \left( \cdot \right)$ are the functions of the BCT, BOT and BBT modules under the $s_t$ sub-distribution, respectively. $q_{s_t}^c, q_{s_t}^o \in \mathbb{R} ^ m$,$q_{s_t}^b \in \mathbb{R} ^ r$, $m$ is the number of buckets divided in the sub-distribution, and $r$ is the number of buckets with bucket width greater than 1 in the sub-distribution, obviously $m \geq r$. $o_{t-1}$ is the normalized bucket multinomial distribution of the upstream task output, which directly monotonically affects the output distribution of the current task after the transformation of the $\emph{Mono}$ function.

The structure of MDME realizes LTV modeling from coarse-grained to fine-grained. Given the user representation vector, the sub-distribution with the highest probability predicted by DCT module is set as the target sub-distribution, and then the BCT module of the target sub-distribution outputs the target bucket with the highest probability. At the same time, we obtain the left boundary and bucket width of the target bucket. Finally, the BBT module outputs the bias coefficient in the target bucket, and we can calculate the estimated LTV value:
\begin{equation}
  u_t = argmax \left( q_{argmax \left( p_t^c \right)}^c \right) 
\end{equation}
\begin{equation}
  \hat{y}_t = {left\_boundary}_{u_t} + q_{argmax \left( p_t^c \right)} ^ b * {bucket\_width}_{u_t}.
\end{equation}

$u_t$ is the bucket number which has the maximum probability in the distribution $q_{argmax( p_t^c )}^c$. It is worth to mention that the MDME module can be disassembled as an independent model to estimate LTV, and still achieves good performance.


\subsection{Normalized Multinomial Distribution and Mono Unit}
The LTVs of different time spans satisfy the ordered relationship, such as $ltv_{30}$ $\le$ $ltv_{90}$ $\le$ $ltv_{180}$ $\le$ $ltv_{365}$, which is determined by business. The conventional modeling strategies are to utilize an independent model to estimate one target, or simply to learn LTVs of multiple time spans at the same time based on multi-task learning. However, they do not make full use of the ordered dependencies between LTVs of different time spans. We believe that modeling this ordered dependencies relationship can effectively improve model performance. Here, through several multi-layer perceptrons with non-negative hidden layer parameters, which we call Mono Unit, the output layers of the upstream and downstream LTV prediction tasks are connected in series, so that the distribution information output by the previous task can monotonically and directly affect the output distribution of the latter task. The reason why Mono Unit connects with the output layer of each task network is that, the closer to the output layer, the richer theoretically the task-related information the hidden layer could get.  \cite{xi2021modeling}.

Specifically, the task needs to output a normalized bucket multinomial distribution. The MDME module performs "coarse-to-fine-grained" processing on the imbalanced LTV distribution. Through distribution segmentation, and sub-distribution bucketing, it greatly reduces the LTV modeling difficulty. These two stages transform LTV modeling into a bucket multi-classification problem, we can calculate the normalized bucket multinomial distribution $o_t$ based on the multinomial distributions output by these two stages.

The normalized bucket multinomial distribution of the output of the $t-1_{th}$ task directly performs the "Add" operation with the output logits of the DCT module and the output logits of the $S_t$ BCT modules respectively after the processing of ($S_t$+1) Mono Units. Mono Unit can capture the shifting trend of the output multinomial distribution of upstream tasks, and then affect the output distribution of downstream tasks. In theory, the modeling difficulty of short-term LTV is less than that of long-term LTV. Through the multinomial distribution mapping and capturing mechanism, short-term LTV can be used to assist the modeling of long-term LTV, and the difficulty of modeling subsequent tasks will be reduced. Meanwhile, it is necessary to do gradient truncation between upstream and downstream tasks, in order to diminish the impact of downstream task on upstream one.

ODMN framework further improves the performance and prediction accuracy by modeling the ordered dependencies of LTVs in different time spans, and achieves soft monotonic constraints. From the perspective of modeling difficulty, the design of Mono Unit enables simple tasks to assist the learning of complex tasks, thereby easing the modeling difficulty of long-term LTV.

\subsection{Fine-grained Calibration and Joint Optimization for MTL}
For each MDME module, the losses associated with DSM module include the sub-distribution multi-classification cross-entropy loss $\mathcal{L}_{t}^c$, the sub-distribution Ordinal Regression loss $\mathcal{L}_{t}^o$, and the sub-distribution distillation classification cross-entropy loss $\mathcal{L}_{t}^{dis}$. The losses of each SDM module consists of the bucket multi-classification cross-entropy loss $\mathcal{L}_{s_t}^c$, the bucket Ordinal Regression loss $\mathcal{L}_{s_t}^o$, the bucket distillation classification cross-entropy loss $\mathcal{L}_{s_t}^{dis}$, and the bucket bias regression loss $\mathcal{L}_{s_t}^b$. The definition of the Ordinal Regression loss is as follows:
\begin{equation}
  \mathcal{L}^o \left( \theta \right) = - \frac{1}{B} \sum_{\left( x, y_t \right) \in \mathcal{D}}^{B} \left( \sum_{u=0}^{y_t-1} log\left( \mathcal{P}^u \right) + \sum_{u=y_t}^{U-1} \left( log \left( 1-\mathcal{P}^u \right)  \right) \right)
\end{equation}

\begin{equation}
\mathcal{P}^u = P\left( \hat{l} > u| \theta \right)
\end{equation}

\begin{equation}
\hat{l} = \sum_{u=0}^{U-1} \eta \left(  \mathcal{P} ^ u \geq 0.5 \right),
\end{equation}
$B$ is the number of samples in a batch, $U$ is the size of the output logits, $y_t$ is the real label corresponding to a sample, and $\hat{l}$ is the estimated value of Ordinal Regression. $\eta$(·) is an indicator function such that $\eta$(true) = 1 and $\eta$(false) = 0, u is bucket number. $\mathcal{P}^u$ is the probability that the predicted ordinal regression value is greater than actual bucket number. It should be noted that the learning of DSM module needs to use all samples, but for the SDM module, only the samples belonging to the sub-distribution or the buckets are utilized to isolate the training samples. Mono Unit captures the shifting trend of the bucket multinomial distribution output by the upstream task, then affects the output distribution of the downstream task, thus realizing soft monotonic constraints in a coarse-grained form. In order to further model the monotonic constraint relationship between upstream and downstream tasks, we perform fine-grained calibration on the estimated LTV values of each task. Specifically, if the short-term LTV estimated by the upstream task is larger than the long-term LTV estimated by the adjacent downstream tasks , a penalty loss is introduced:
\begin{equation}
\mathcal{L}_t^{cali} =  \frac{1}{B} \sum_{\left( x, y_t \right) \in \mathcal{D}}^{B} \sum_{t=1}^{T-1} max\left( 
\hat{y}_t - \hat{y}_{t+1} , 0\right).
\end{equation}

In summary, the loss function of ODMN is defined as follows, $\alpha$ controls the strength of the fine-grained ordered dependency calibration loss, $\beta$ and $\gamma_{s_t}$ determines the contribution of each distillation loss:
\begin{equation}
\mathcal{L} = \sum_{t=1}^{T} \mathcal{L}_t + \alpha * \mathcal{L}_t^{cali}
\end{equation}
\begin{equation}
\mathcal{L}_t = \mathcal{L}_t^c + \mathcal{L}_t^o + \beta * \mathcal{L}_{t}^{dis} +  \sum_{s_t=1}^{S_t} \left( \mathcal{L}_{s_t}^c + \mathcal{L}_{s_t}^o + \gamma_{s_t} * \mathcal{L}_{s_t}^{dis} + \mathcal{L}_{s_t}^b \right).
\end{equation}

\section{Evaluation Metrics}
\subsection{Common Evaluation Metrics}
\begin{figure}[h]
  \centering
  \includegraphics[width=\linewidth]{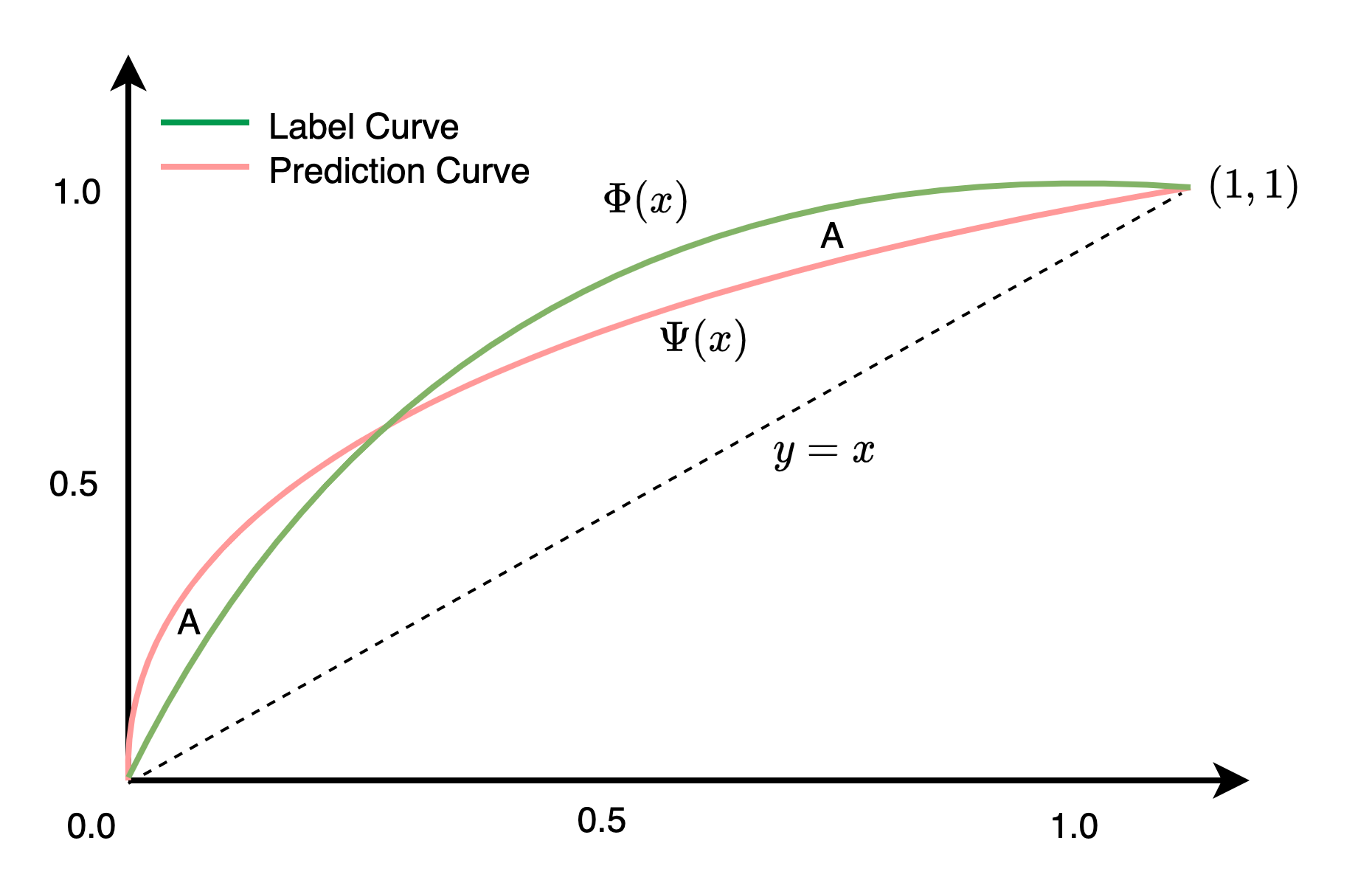}
  \caption{Lorenz Curve of real label and model prediction, the horizontal axis is the cumulative percentage of the number of users, and the vertical axis is the cumulative percentage of LTV contributed by users. Note that the original definition was to sort the true LTV in ascending order, we change it for more straightforward interpretations of high-value customers by sorting the true LTV in descending order.}
  \Description{Lorenz Curve of the model}
  \label{fig:lorenz}
\end{figure}

We adopt Normalized Rooted Mean Square Error (NRMSE) and Normalized Mean Average Error (NMAE) defined in \cite{xing2021learning} as parts of the evaluation metrics. The absolute value of Mean Bias Error (AMBE) is another important metric, which can capture the average bias in the prediction. Note that, the lower these three metrics are, the performance is better. However, the above metrics can not measure the model's ability to differentiate high-value users from low-value users, nor can they reflect how well the model fits the real imbalanced distribution. 
\newline
\textbf{ZILN} \cite{wang2019deep} introduce the normalized model Gini based on the Lorenz Curve and the Gini Coefficient. Among them, the Lorenz Curve (Figure \ref{fig:lorenz}) is an intuitive tool to describe the distribution of high and low value users. The horizontal axis of the Lorenz Curve is the cumulative percentage of the number of users, and the vertical axis is the cumulative percentage of LTV contributed by users. The fit degree of the model estimated value curve and the true value curve reflects the fitting ability of the model to the imbalanced distribution. Gini Coefficient is a discriminative measure. The larger the 
estimated Gini Coefficient, the better the model's ability of discrimination. For an imbalanced dataset, assuming that the total sample size is $N$, $N_i$ is the sample size of the class $i$, and $C$ is the total class size, then Gini Coefficient can be defined as:
\begin{equation}
I_{Gini}=\frac {\sum_{i=1}^C \left( 2i-C-1 \right)  N_{i}}{C\sum_{i=1}^C N_{i}}.
\end{equation}
The ratio between the model Gini coefficient and the true Gini coefficient yields the normalized model Gini coefficient which measure fit degree of the estimated value curve and the true value curve. However, the normalized model Gini coefficient ignores that curves might crossover. As shown in Figure \ref{fig:lorenz}, when the ratio is 1, the curves might not be overlapped.

\subsection{Mutual Gini}
\label{section:Mutual_Gini}
In this paper, we propose a new evaluation metric called $\emph{Mutual}$ $\emph{Gini}$, which can quantitatively measure the difference between the curves based on the Lorenz Curve. As shown in Figure \ref{fig:lorenz}, the green curve is the Lorenz Curve of the real label, and the red one belongs to the estimated value, the $\emph{Mutual}$ $\emph{Gini}$ is defined as the area $A$ between the green curve and the red one. The smaller the $\emph{Mutual}$ $\emph{Gini}$, the better the model fits the real imbalanced distribution. The calculation of $\emph{Mutual}$ $\emph{Gini}$ is as follows:
\begin{equation}
\begin{split}
    Mutual\_Gini = \int_{0}^{1} \left |\Phi(x)-\Psi(x)  \right | dx,
\end{split}
\end{equation}
$\Phi(x)$ and $\Psi(x)$ are the Lorenz Curves of the true label and the estimated value respectively. As we can see, the definition of $\emph{Mutual}$ $\emph{Gini}$ helps us to measure the overall accuracy of the LTV distribution, rather than the point-wise loss of LTV prediction. This is very critical as we usually rely on this distribution to make real-world operation decisions.

\section{Experiments}
In this section, we perform experiments to evaluate the proposed framework against baseline models on real-world industrial data. We firstly introduce the datasets and experiments setup. Besides that, a novel evaluation metric named \emph{Mutual Gini} is introduced, which we think can accurately measure the model's performance to fit real imbalanced distribution. Finally, the experimental results and analysis will be presented.

\subsection{Experiments Setup}

\subsubsection{Datasets}
The dataset of this experiment is from Kuaishou's user growth business. Kuaishou is a company whose main business is mobile short videos. As of November 2021, the DAU of Kuaishou reached 320 million and MAU has exceeded 1 billion. We sample 180 million new users as the experimental datasets, in which, the data were collected from multiple user acquisition channels, like pre-installation on new mobile devices, app store download, fission and so on. The features of the dataset consist of user profile data, channel-related information, and the behavior information of new users within 7 days after registering the Kuaishou's application. Besides that, we also collect some periodic, seasonal or abnormal event information as parts of the features. The features are divided into categorical type and numerical type, and the label is defined as the active days for new users in the next month, quarter, half a year, and one year after registering the Kuaishou's application. That is, based on the behavior data and user attributes of new users in the first 7 days, we need estimate $ltv_{30}$, $ltv_{90}$, $ltv_{180}$, $ltv_{365}$ at the same time. As shown in Figure \ref{fig:data}, is the $ltv_{365}$ distribution of the dataset.

\subsubsection{Experiments Setup}
To verify the performance of the model, we evaluate the model's prediction of long and short-term LTV (e.g. $ltv_{30}$, $ltv_{90}$, $ltv_{180}$, $ltv_{365}$). The optimizer we use is Adam, the batch size is empirically set to 512, the learning rate of the network parameters is 0.05, and the embeddings' learning rate is set to 0.1. The ordered dependency calibration loss weight $\alpha$ is set as 1, and the Ordinal Regression distillation loss weight $\beta$ and $\gamma$ are both set to 0.5. In all MDME modules, the number of sub-distributions is 2, and the cut point depends on the specific distribution. The number of buckets for each sub-distribution also depends on how imbalanced the sub-distributions are. Empirically, the number of sub-distributions and the number of buckets in each sub-distribution have an effect on the final performance, but not very much.

\subsection{Performance Comparison}
Here, we choose two well-known methods as baselines, namely (1) Two-Stage \cite{drachen2018or}, (2) ZILN \cite{wang2019deep}. The reason why TSUR \cite{xing2021learning} and other approaches are not utilized as baselines is that, they focus on user representation learning, which can be compatible with our framework for better model performance.   

\begin{itemize}
    \item \textbf{Two-Stage} \cite{drachen2018or} uses a two-step process to predict LTV, it first classifies whether a user was premium or not, followed by predicting the monetary value that the user brings. 
    \item \textbf{ZILN} \cite{wang2019deep} assumes the zero-inflated long-tailed LTV obeys the log-normal distribution, and takes advantages of DNN in feature intersection and information capture to fit the mean and standard deviation of the distribution, and finally the expected value of the log-normal distribution is used as the estimated LTV. 
\end{itemize}

Table \ref{tab:table1} presents the results of all the baselines and our model on $ltv_{30}$, $ltv_{90}$, $ltv_{180}$, $ltv_{365}$ prediction. Firstly, Two-Stage method is the worst on every evaluation metric. Secondly, ZILN works better, but its performance on the AMBE is poor, and both ZILN and Two-Stage methods performs poorly on \textit{Mutual} \textit{Gini}, which means that they can not fit the imbalanced LTV distribution well, as can be seen from the Gini Coefficient, although we do not think the Gini Coefficient to be a good metric of model performance. Finally, compared to the above two methods, ODMN has a significant improvement in all metrics, which shows the best performance.

\begin{table}
  \caption{Performance comparison on next 30-day 90-day 180-day and 365-day LTV prediction. Best results are marked in bold. Gini Coefficient is not a good choice to reflect the performance difference of fitting imbalanced distribution between the models very well, but we still present it.}
  \label{tab:table1}
  \scalebox{0.85}{
  \begin{tabular}{ccccccl}
    \toprule
    METHORD          & NRMSE & NMAE & AMBE & Mutual Gini & $Gini^{*}$\\
    \midrule
    $Two-Stage_{30}$\cite{drachen2018or} & 0.8888	& 0.5768 & 0.3181 & 0.0273 & 0.4737 \\
    $ZILN_{30}$\cite{wang2019deep}	     & 0.4831 & 0.2963 & 0.1336 & 0.0226 & 0.4996 \\
    $ODMN_{30}$	     & \textbf{0.4765} & \textbf{0.2683} & \textbf{0.0423} & \textbf{0.0125} & 0.5688 \\
    \midrule
    $Two-Stage_{90}$\cite{drachen2018or} & 1.1554 & 0.8159 & 1.8596 & 0.0550 & 0.5067 \\
    $ZILN_{90}$\cite{wang2019deep}      & 0.7759 & 0.4707 & 1.4697 & 0.0470 & 0.5318 \\
    $ODMN_{90}$	     & \textbf{0.7514} & \textbf{0.4158} & \textbf{0.1263} & \textbf{0.0169} & 0.6617 \\
	\midrule				
    $Two-Stage_{180}$\cite{drachen2018or} & 1.3992 & 0.9791 & 3.4375 & 0.0639 & 0.5386 \\
    $ZILN_{180}$\cite{wang2019deep}	 & 0.9441 & 0.5761 & 4.9021 & 0.0543 & 0.5742 \\
    $ODMN_{180}$	 & \textbf{0.9037} & \textbf{0.4812} & \textbf{0.2460} & \textbf{0.0098} & 0.7054 \\
    \midrule
    $Two-Stage_{365}$\cite{drachen2018or} & 1.7507 & 1.1749 & 6.6276 & 0.0821 & 0.5727 \\
    $ZILN_{365}$\cite{wang2019deep}	 & 1.2103 & 0.6833 & 8.6644 & 0.0558 & 0.6251 \\
    $ODMN_{365}$	 & \textbf{1.1538} & \textbf{0.5833} & \textbf{0.3123} & \textbf{0.0083} & 0.7507 \\
    \bottomrule
  \end{tabular}
}
\end{table}

\subsection{Ablation Study}
In this section, we conduct ablation experiments to verify the impact of different modules in our model. The experiments are carried out in two directions, one is the single time granularity LTV modeling for the purpose of MDME module validation, and the other is the validation of ODMN framework to model multi-time-span LTVs.

\subsubsection{MDME}
Since $ltv_{365}$ is more difficult to model, we examine the effects of the core modules or structures in MDME by predicting $ltv_{365}$. Four variants of MDME are compared, including: \textbf{(A)} \underline{NM} denotes \textit{Naive MDME}, relying only on the distribution segmentation and bucketing mechanism, without bucket bias regression and Ordinal Regression distillation, the mean of the left and right boundaries of the estimated bucket is taken as the final estimated LTV. \textbf{(B)} \underline{NMB} denotes combining \textit{Naive MDME} and bucket bias regression to predict LTV. \textbf{(C)} \underline{NMO} denotes combining \textit{Naive MDME} and Ordinal Regression distillation to predict LTV. \textbf{(D)} \underline{MDME} denotes the complete module we design.

\begin{table}
  \caption{The ablation study on four variants of MDME. Best results are marked in bold.}
  \label{tab:table2}
  \scalebox{0.95}{
  \begin{tabular}{ccccccl}
    \toprule
    METHORD          & NRMSE & NMAE & AMBE  & Mutual Gini & $Gini^{*}$\\
    \midrule
    $NM_{365}$       & 1.4129 & 0.6795 & 8.3200 & 0.0194 &	0.7626\\
    $NMB_{365}$	     & 1.4006 & 0.6707 & 5.3401 & 0.0201 &	0.7707\\
    $NMO_{365}$      & \textbf{1.2064} & \textbf{0.6006} & 4.1688 & 0.0151 & 0.7669\\
    $MDME_{365}$     & 1.2119 & 0.6081 & \textbf{0.5606} & \textbf{0.0127} & 0.7622\\
  \bottomrule
\end{tabular}
}
\end{table}

In Table \ref{tab:table2}, the performance ranking of each variant can be roughly summarized as: \underline{NM} < \underline{NMB} < \underline{NMO} < \underline{MDME}. We can infer three core conclusions: (1). The introduction of BBT within each sub-distribution reduces mean bias to a certain extent. (2). Through the auxiliary distillation of Ordinal Regression, the metrics such as \textit{Mutual} \textit{Gini}, NRMSE and NMAE are greatly reduced, indicating that the model is better at learning the ordered relationship between sub-distributions and between buckets. (3). MDME shows the best performance overall.

\subsubsection{ODMN}
We can verify the utility of the core modules in ODMN by simultaneously estimating $ltv_{30}$, $ltv_{90}$, $ltv_{180}$ and $ltv_{365}$. There are four variants: \textbf{(E)} \underline{S} denotes using the conventional share-bottom multi-task learning to predict multiple LTV targets, with each task modeling by MDME. \textbf{(F)} \underline{SM} denotes combining \underline{S} and \textit{Mono Unit} to predict multiple LTV targets. \textbf{(G)} \underline{SC} denotes combining \underline{S} and ordered dependency calibration loss to predict multiple LTV targets. \textbf{(H)} \underline{ODMN} denotes the complete framework we propose.

\begin{table}
  \caption{The ablation study on four variants of ODMN. Best results are marked in bold.}
  \label{tab:table3}
  \scalebox{0.95}{
  \begin{tabular}{ccccccl}
    \toprule
    METHORD          & NRMSE &NMAE & AMBE & Mutual Gini & $Gini^{*}$\\
    \midrule
    $S_{30}$     & 0.4877 & 0.2765 & 0.1098  & 0.0147 & 0.5591\\
    $SM_{30}$	 & 0.4860 & 0.2726 & 0.0678  & 0.0131 & 0.5687\\
    $SC_{30}$    & 0.4816 & 0.2704 & 0.0744  & 0.0134 & 0.5631\\
    $ODMN_{30}$  & \textbf{0.4765} & \textbf{0.2683} & \textbf{0.0423} & \textbf{0.0125} & 0.5688\\
    \midrule
    $S_{90}$       & 0.7784 & 0.4338 & 0.3787  & 0.0215 & 0.6685\\
    $SM_{90}$	   & 0.7620 & 0.4187 & 0.2161  & 0.0185 & 0.6589\\
    $SC_{90}$      & 0.7606 & 0.4183 & 0.1528  & 0.0193 & 0.6623\\
    $ODMN_{90}$    & \textbf{0.7514} & \textbf{0.4158} & \textbf{0.1263} & \textbf{0.0169} & 0.6617\\
    \midrule
    $S_{180}$       & 0.9484 & 0.5013 & 0.5787  & 0.0135 & 0.7096\\
    $SM_{180}$	    & 0.9246 & 0.4893 & 0.3561  & 0.0121 & 0.7071\\
    $SC_{180}$      & 0.9281 & 0.4913 & 0.3521  & 0.0113 & 0.7134\\
    $ODMN_{180}$    & \textbf{0.9037} & \textbf{0.4812} & \textbf{0.2460} & \textbf{0.0098} & 0.7054\\
    \midrule
    $S_{365}$       & 1.2163 & 0.6086 & 0.6475  & 0.0144 & 0.7305\\
    $SM_{365}$	    & 1.1995 & 0.6057 & 0.5496  & 0.0092 & 0.7497\\
    $SC_{365}$      & 1.1862 & 0.6002 & 0.4155  & 0.0101 & 0.7568\\
    $ODMN_{365}$    & \textbf{1.1538} & \textbf{0.5833} & \textbf{0.3123} & \textbf{0.0083} & 0.7507\\
  \bottomrule
\end{tabular}
}
\end{table}

Table \ref{tab:table3} presents the performance of four variants of ODMN. Conventional multi-task learning has limited improvement. \textit{Mono} \textit{Unit} and its associated structures really play an important role in capturing the monotonic constraints between long and short-term LTVs. The fine-grained ordered dependency calibration loss further enhances this kind of monotonic constraints. In addition, there are some facts that are in line with expectations. From $ltv_{30}$ to $ltv_{365}$, the increase of Gini Coefficient shows that the imbalance degree of LTV distribution increases, and the modeling difficulty also increases accordingly. However, compared with short-term LTV, long-term LTV has more space for improvement, indicating that short-term LTV can indeed assist long-term LTV modeling in ODMN, which proves the effectiveness of the framework.

\subsection{Online A/B Test}
To further verify the effectiveness of the model, we conduct A/B tests in the advertising for Kuaishou user growth. Specifically, we take 10$\%$ of the traffic as the control group and 10$\%$ of the traffic as the experimental group, use different models to estimate the $ltv_{30}$ of the users, and deliver ads to users with highest ROI. For the convenience of observing the experimental results, the modeling label in $ltv_{30}$ estimation is set as the value that the user will bring to the platform after 30 days. The delivery strategy afterwards is the same (omitted to company privacy) to the two groups. We accumulate experimental data for 7 days, 14 days, and 30 days, and calculate the ROI (Return on Investment, omitted to company privacy) for both groups. In terms of model, we choose ZILN with the state-of-the-art performance as the baseline, and use our method as the model of the experimental group.

\begin{table}
  \caption{Comparison on ROI metrics in Online A/B test.}
  \label{tab:table4}
  \scalebox{0.95}{
  \begin{tabular}{ccccccl}
    \toprule
    METHORD     & ROI-7 & ROI-14 & ROI-30 \\
    \midrule
    ROI UPLIFT  & 11.9$\%$ & 12.8$\%$ & 14.7$\%$  \\
  \bottomrule
\end{tabular}
}
\end{table}

For company privacy, only the ROI improvement value of our method relative to the baseline is provided in Table \ref{tab:table4}. it is obvious that our method performs better, which further demonstrates the effectiveness of the proposed model.

Our model has been fully deployed in the Kuaishou user growth business. We adopts the method of day-level training and prediction of full volume users. Then, we save the user's estimated LTV in the cache for online real-time acquisition.

\section{Conclusion}
In this paper, we proposed an Order Dependency Monotonic Network for LTV modeling, which made full use of the ordered dependencies between long-term and short-term LTVs. By capturing the shifting trend of the bucket multinomial distribution of the short-term LTV output, and directly affecting the long-term LTV output distribution, we achieved soft monotonic constraints in a coarse-grained manner. In addition, the utilization of the ordered dependency calibration loss further enhanced the monotonic constraint relationship. For the modeling of a specific LTV, we proposed a novel MDME module based on the idea of \textit{Divide-and-Conquer}, which divided the imbalanced LTV distribution step by step to alleviate the imbalanced degree. The Ordinal Regression assisted distillation further improved the ranking and classification accuracy between sub-distributions and between buckets, which greatly improved the performance of the model. Finally, in order to accurately measure the model's ability to fit real imbalanced distribution, we proposed the metric of \emph{Mutual} \emph{Gini} on basis of Lorenz Curve. Our approach achieved considerable gains in both offline experiments on real-world industrial datasets and online application, demonstrating the effectiveness of the approach.


\bibliographystyle{ACM-Reference-Format}
\balance
\bibliography{main}


\begin{thebibliography}{23}


\ifx \showCODEN    \undefined \def \showCODEN     #1{\unskip}     \fi
\ifx \showDOI      \undefined \def \showDOI       #1{#1}\fi
\ifx \showISBNx    \undefined \def \showISBNx     #1{\unskip}     \fi
\ifx \showISBNxiii \undefined \def \showISBNxiii  #1{\unskip}     \fi
\ifx \showISSN     \undefined \def \showISSN      #1{\unskip}     \fi
\ifx \showLCCN     \undefined \def \showLCCN      #1{\unskip}     \fi
\ifx \shownote     \undefined \def \shownote      #1{#1}          \fi
\ifx \showarticletitle \undefined \def \showarticletitle #1{#1}   \fi
\ifx \showURL      \undefined \def \showURL       {\relax}        \fi
\providecommand\bibfield[2]{#2}
\providecommand\bibinfo[2]{#2}
\providecommand\natexlab[1]{#1}
\providecommand\showeprint[2][]{arXiv:#2}

\bibitem[\protect\citeauthoryear{Chang, Chang, and Li}{Chang
  et~al\mbox{.}}{2012}]%
        {chang2012customer}
\bibfield{author}{\bibinfo{person}{Wen Chang}, \bibinfo{person}{Chen Chang},
  {and} \bibinfo{person}{Qianpin Li}.} \bibinfo{year}{2012}\natexlab{}.
\newblock \showarticletitle{Customer lifetime value: A review}.
\newblock \bibinfo{journal}{\emph{Social Behavior and Personality: an
  international journal}} \bibinfo{volume}{40}, \bibinfo{number}{7}
  (\bibinfo{year}{2012}), \bibinfo{pages}{1057--1064}.
\newblock


\bibitem[\protect\citeauthoryear{Chen, Guitart, del R{\'\i}o, and
  Peri{\'a}nez}{Chen et~al\mbox{.}}{2018}]%
        {chen2018customer}
\bibfield{author}{\bibinfo{person}{Pei~Pei Chen}, \bibinfo{person}{Anna
  Guitart}, \bibinfo{person}{Ana~Fern{\'a}ndez del R{\'\i}o}, {and}
  \bibinfo{person}{Africa Peri{\'a}nez}.} \bibinfo{year}{2018}\natexlab{}.
\newblock \showarticletitle{Customer lifetime value in video games using deep
  learning and parametric models}. In \bibinfo{booktitle}{\emph{2018 IEEE
  international conference on big data (big data)}}. IEEE,
  \bibinfo{pages}{2134--2140}.
\newblock


\bibitem[\protect\citeauthoryear{Devlin, Chang, Lee, and Toutanova}{Devlin
  et~al\mbox{.}}{2018}]%
        {devlin2018bert}
\bibfield{author}{\bibinfo{person}{Jacob Devlin}, \bibinfo{person}{Ming-Wei
  Chang}, \bibinfo{person}{Kenton Lee}, {and} \bibinfo{person}{Kristina
  Toutanova}.} \bibinfo{year}{2018}\natexlab{}.
\newblock \showarticletitle{Bert: Pre-training of deep bidirectional
  transformers for language understanding}.
\newblock \bibinfo{journal}{\emph{arXiv preprint arXiv:1810.04805}}
  (\bibinfo{year}{2018}).
\newblock


\bibitem[\protect\citeauthoryear{Drachen, Pastor, Liu, Fontaine, Chang, Runge,
  Sifa, and Klabjan}{Drachen et~al\mbox{.}}{2018}]%
        {drachen2018or}
\bibfield{author}{\bibinfo{person}{Anders Drachen}, \bibinfo{person}{Mari
  Pastor}, \bibinfo{person}{Aron Liu}, \bibinfo{person}{Dylan~Jack Fontaine},
  \bibinfo{person}{Yuan Chang}, \bibinfo{person}{Julian Runge},
  \bibinfo{person}{Rafet Sifa}, {and} \bibinfo{person}{Diego Klabjan}.}
  \bibinfo{year}{2018}\natexlab{}.
\newblock \showarticletitle{To be or not to be... social: Incorporating simple
  social features in mobile game customer lifetime value predictions}. In
  \bibinfo{booktitle}{\emph{Proceedings of the Australasian Computer Science
  Week Multiconference}}. \bibinfo{pages}{1--10}.
\newblock


\bibitem[\protect\citeauthoryear{Erhan, Courville, Bengio, and Vincent}{Erhan
  et~al\mbox{.}}{2010}]%
        {erhan2010does}
\bibfield{author}{\bibinfo{person}{Dumitru Erhan}, \bibinfo{person}{Aaron
  Courville}, \bibinfo{person}{Yoshua Bengio}, {and} \bibinfo{person}{Pascal
  Vincent}.} \bibinfo{year}{2010}\natexlab{}.
\newblock \showarticletitle{Why does unsupervised pre-training help deep
  learning?}. In \bibinfo{booktitle}{\emph{Proceedings of the thirteenth
  international conference on artificial intelligence and statistics}}. JMLR
  Workshop and Conference Proceedings, \bibinfo{pages}{201--208}.
\newblock


\bibitem[\protect\citeauthoryear{Fader, Hardie, and Lee}{Fader
  et~al\mbox{.}}{2005}]%
        {fader2005rfm}
\bibfield{author}{\bibinfo{person}{Peter~S Fader}, \bibinfo{person}{Bruce~GS
  Hardie}, {and} \bibinfo{person}{Ka~Lok Lee}.}
  \bibinfo{year}{2005}\natexlab{}.
\newblock \showarticletitle{RFM and CLV: Using iso-value curves for customer
  base analysis}.
\newblock \bibinfo{journal}{\emph{Journal of marketing research}}
  \bibinfo{volume}{42}, \bibinfo{number}{4} (\bibinfo{year}{2005}),
  \bibinfo{pages}{415--430}.
\newblock


\bibitem[\protect\citeauthoryear{Fu, Gong, Wang, Batmanghelich, and Tao}{Fu
  et~al\mbox{.}}{2018}]%
        {fu2018deep}
\bibfield{author}{\bibinfo{person}{Huan Fu}, \bibinfo{person}{Mingming Gong},
  \bibinfo{person}{Chaohui Wang}, \bibinfo{person}{Kayhan Batmanghelich}, {and}
  \bibinfo{person}{Dacheng Tao}.} \bibinfo{year}{2018}\natexlab{}.
\newblock \showarticletitle{Deep ordinal regression network for monocular depth
  estimation}. In \bibinfo{booktitle}{\emph{Proceedings of the IEEE conference
  on computer vision and pattern recognition}}. \bibinfo{pages}{2002--2011}.
\newblock


\bibitem[\protect\citeauthoryear{Guo, Tang, Ye, Li, and He}{Guo
  et~al\mbox{.}}{2017}]%
        {guo2017deepfm}
\bibfield{author}{\bibinfo{person}{Huifeng Guo}, \bibinfo{person}{Ruiming
  Tang}, \bibinfo{person}{Yunming Ye}, \bibinfo{person}{Zhenguo Li}, {and}
  \bibinfo{person}{Xiuqiang He}.} \bibinfo{year}{2017}\natexlab{}.
\newblock \showarticletitle{DeepFM: a factorization-machine based neural
  network for CTR prediction}.
\newblock \bibinfo{journal}{\emph{arXiv preprint arXiv:1703.04247}}
  (\bibinfo{year}{2017}).
\newblock


\bibitem[\protect\citeauthoryear{Gupta, Hanssens, Hardie, Kahn, Kumar, Lin,
  Ravishanker, and Sriram}{Gupta et~al\mbox{.}}{2006}]%
        {gupta2006modeling}
\bibfield{author}{\bibinfo{person}{Sunil Gupta}, \bibinfo{person}{Dominique
  Hanssens}, \bibinfo{person}{Bruce Hardie}, \bibinfo{person}{Wiliam Kahn},
  \bibinfo{person}{V Kumar}, \bibinfo{person}{Nathaniel Lin},
  \bibinfo{person}{Nalini Ravishanker}, {and} \bibinfo{person}{S Sriram}.}
  \bibinfo{year}{2006}\natexlab{}.
\newblock \showarticletitle{Modeling customer lifetime value}.
\newblock \bibinfo{journal}{\emph{Journal of service research}}
  \bibinfo{volume}{9}, \bibinfo{number}{2} (\bibinfo{year}{2006}),
  \bibinfo{pages}{139--155}.
\newblock


\bibitem[\protect\citeauthoryear{He, Wu, and Wei}{He et~al\mbox{.}}{2021}]%
        {he2021distilling}
\bibfield{author}{\bibinfo{person}{Yin-Yin He}, \bibinfo{person}{Jianxin Wu},
  {and} \bibinfo{person}{Xiu-Shen Wei}.} \bibinfo{year}{2021}\natexlab{}.
\newblock \showarticletitle{Distilling Virtual Examples for Long-tailed
  Recognition}.
\newblock \bibinfo{journal}{\emph{arXiv preprint arXiv:2103.15042}}
  (\bibinfo{year}{2021}).
\newblock


\bibitem[\protect\citeauthoryear{Ma, Zhao, Yi, Chen, Hong, and Chi}{Ma
  et~al\mbox{.}}{2018}]%
        {ma2018modeling}
\bibfield{author}{\bibinfo{person}{Jiaqi Ma}, \bibinfo{person}{Zhe Zhao},
  \bibinfo{person}{Xinyang Yi}, \bibinfo{person}{Jilin Chen},
  \bibinfo{person}{Lichan Hong}, {and} \bibinfo{person}{Ed~H Chi}.}
  \bibinfo{year}{2018}\natexlab{}.
\newblock \showarticletitle{Modeling task relationships in multi-task learning
  with multi-gate mixture-of-experts}. In \bibinfo{booktitle}{\emph{Proceedings
  of the 24th ACM SIGKDD International Conference on Knowledge Discovery \&
  Data Mining}}. \bibinfo{pages}{1930--1939}.
\newblock


\bibitem[\protect\citeauthoryear{Talaba et~al\mbox{.}}{Talaba
  et~al\mbox{.}}{2013}]%
        {talaba2013comparison}
\bibfield{author}{\bibinfo{person}{Monica Talaba} {et~al\mbox{.}}}
  \bibinfo{year}{2013}\natexlab{}.
\newblock \showarticletitle{Comparison between Customer Lifetime Value (CLV)
  and traditional measurement tools of customer value}.
\newblock \bibinfo{journal}{\emph{Knowledge Horizons-Economics}}
  \bibinfo{volume}{5}, \bibinfo{number}{2} (\bibinfo{year}{2013}),
  \bibinfo{pages}{189--192}.
\newblock


\bibitem[\protect\citeauthoryear{Tang, Liu, Zhao, and Gong}{Tang
  et~al\mbox{.}}{2020}]%
        {tang2020progressive}
\bibfield{author}{\bibinfo{person}{Hongyan Tang}, \bibinfo{person}{Junning
  Liu}, \bibinfo{person}{Ming Zhao}, {and} \bibinfo{person}{Xudong Gong}.}
  \bibinfo{year}{2020}\natexlab{}.
\newblock \showarticletitle{Progressive layered extraction (ple): A novel
  multi-task learning (mtl) model for personalized recommendations}. In
  \bibinfo{booktitle}{\emph{Fourteenth ACM Conference on Recommender Systems}}.
  \bibinfo{pages}{269--278}.
\newblock


\bibitem[\protect\citeauthoryear{Vanderveld, Pandey, Han, and
  Parekh}{Vanderveld et~al\mbox{.}}{2016}]%
        {vanderveld2016engagement}
\bibfield{author}{\bibinfo{person}{Ali Vanderveld}, \bibinfo{person}{Addhyan
  Pandey}, \bibinfo{person}{Angela Han}, {and} \bibinfo{person}{Rajesh
  Parekh}.} \bibinfo{year}{2016}\natexlab{}.
\newblock \showarticletitle{An engagement-based customer lifetime value system
  for e-commerce}. In \bibinfo{booktitle}{\emph{Proceedings of the 22nd ACM
  SIGKDD international conference on knowledge discovery and data mining}}.
  \bibinfo{pages}{293--302}.
\newblock


\bibitem[\protect\citeauthoryear{Vaswani, Shazeer, Parmar, Uszkoreit, Jones,
  Gomez, Kaiser, and Polosukhin}{Vaswani et~al\mbox{.}}{2017}]%
        {vaswani2017attention}
\bibfield{author}{\bibinfo{person}{Ashish Vaswani}, \bibinfo{person}{Noam
  Shazeer}, \bibinfo{person}{Niki Parmar}, \bibinfo{person}{Jakob Uszkoreit},
  \bibinfo{person}{Llion Jones}, \bibinfo{person}{Aidan~N Gomez},
  \bibinfo{person}{{\L}ukasz Kaiser}, {and} \bibinfo{person}{Illia
  Polosukhin}.} \bibinfo{year}{2017}\natexlab{}.
\newblock \showarticletitle{Attention is all you need}. In
  \bibinfo{booktitle}{\emph{Advances in neural information processing
  systems}}. \bibinfo{pages}{5998--6008}.
\newblock


\bibitem[\protect\citeauthoryear{Wadsworth}{Wadsworth}{2012}]%
        {wadsworth2012buy}
\bibfield{author}{\bibinfo{person}{Edward Wadsworth}.}
  \bibinfo{year}{2012}\natexlab{}.
\newblock \bibinfo{title}{Buy’Til You Die-A Walkthrough}.
\newblock
\newblock


\bibitem[\protect\citeauthoryear{Wang, Fu, Fu, and Wang}{Wang
  et~al\mbox{.}}{2017}]%
        {wang2017deep}
\bibfield{author}{\bibinfo{person}{Ruoxi Wang}, \bibinfo{person}{Bin Fu},
  \bibinfo{person}{Gang Fu}, {and} \bibinfo{person}{Mingliang Wang}.}
  \bibinfo{year}{2017}\natexlab{}.
\newblock \showarticletitle{Deep \& cross network for ad click predictions}.
\newblock In \bibinfo{booktitle}{\emph{Proceedings of the ADKDD'17}}.
  \bibinfo{pages}{1--7}.
\newblock


\bibitem[\protect\citeauthoryear{Wang, Liu, and Miao}{Wang
  et~al\mbox{.}}{2019}]%
        {wang2019deep}
\bibfield{author}{\bibinfo{person}{Xiaojing Wang}, \bibinfo{person}{Tianqi
  Liu}, {and} \bibinfo{person}{Jingang Miao}.} \bibinfo{year}{2019}\natexlab{}.
\newblock \showarticletitle{A deep probabilistic model for customer lifetime
  value prediction}.
\newblock \bibinfo{journal}{\emph{arXiv preprint arXiv:1912.07753}}
  (\bibinfo{year}{2019}).
\newblock


\bibitem[\protect\citeauthoryear{Xi, Chen, Yan, Zhang, Zhu, Zhuang, and
  Chen}{Xi et~al\mbox{.}}{2021}]%
        {xi2021modeling}
\bibfield{author}{\bibinfo{person}{Dongbo Xi}, \bibinfo{person}{Zhen Chen},
  \bibinfo{person}{Peng Yan}, \bibinfo{person}{Yinger Zhang},
  \bibinfo{person}{Yongchun Zhu}, \bibinfo{person}{Fuzhen Zhuang}, {and}
  \bibinfo{person}{Yu Chen}.} \bibinfo{year}{2021}\natexlab{}.
\newblock \showarticletitle{Modeling the Sequential Dependence among Audience
  Multi-step Conversions with Multi-task Learning in Targeted Display
  Advertising}.
\newblock \bibinfo{journal}{\emph{arXiv preprint arXiv:2105.08489}}
  (\bibinfo{year}{2021}).
\newblock


\bibitem[\protect\citeauthoryear{Xiang, Ding, and Han}{Xiang
  et~al\mbox{.}}{2020}]%
        {xiang2020learning}
\bibfield{author}{\bibinfo{person}{Liuyu Xiang}, \bibinfo{person}{Guiguang
  Ding}, {and} \bibinfo{person}{Jungong Han}.} \bibinfo{year}{2020}\natexlab{}.
\newblock \showarticletitle{Learning from multiple experts: Self-paced
  knowledge distillation for long-tailed classification}. In
  \bibinfo{booktitle}{\emph{European Conference on Computer Vision}}. Springer,
  \bibinfo{pages}{247--263}.
\newblock


\bibitem[\protect\citeauthoryear{Xing, Bian, Zhao, Xiao, Luo, Yin, Cai, and
  He}{Xing et~al\mbox{.}}{2021}]%
        {xing2021learning}
\bibfield{author}{\bibinfo{person}{Mingzhe Xing}, \bibinfo{person}{Shuqing
  Bian}, \bibinfo{person}{Wayne~Xin Zhao}, \bibinfo{person}{Zhen Xiao},
  \bibinfo{person}{Xinji Luo}, \bibinfo{person}{Cunxiang Yin},
  \bibinfo{person}{Jing Cai}, {and} \bibinfo{person}{Yancheng He}.}
  \bibinfo{year}{2021}\natexlab{}.
\newblock \showarticletitle{Learning Reliable User Representations from
  Volatile and Sparse Data to Accurately Predict Customer Lifetime Value}. In
  \bibinfo{booktitle}{\emph{Proceedings of the 27th ACM SIGKDD Conference on
  Knowledge Discovery \& Data Mining}}. \bibinfo{pages}{3806--3816}.
\newblock


\bibitem[\protect\citeauthoryear{Yin, Yu, Sohn, Liu, and Chandraker}{Yin
  et~al\mbox{.}}{2019}]%
        {yin2019feature}
\bibfield{author}{\bibinfo{person}{Xi Yin}, \bibinfo{person}{Xiang Yu},
  \bibinfo{person}{Kihyuk Sohn}, \bibinfo{person}{Xiaoming Liu}, {and}
  \bibinfo{person}{Manmohan Chandraker}.} \bibinfo{year}{2019}\natexlab{}.
\newblock \showarticletitle{Feature transfer learning for face recognition with
  under-represented data}. In \bibinfo{booktitle}{\emph{Proceedings of the
  IEEE/CVF Conference on Computer Vision and Pattern Recognition}}.
  \bibinfo{pages}{5704--5713}.
\newblock


\bibitem[\protect\citeauthoryear{Zhang, Kang, Hooi, Yan, and Feng}{Zhang
  et~al\mbox{.}}{2021}]%
        {zhang2021deep}
\bibfield{author}{\bibinfo{person}{Yifan Zhang}, \bibinfo{person}{Bingyi Kang},
  \bibinfo{person}{Bryan Hooi}, \bibinfo{person}{Shuicheng Yan}, {and}
  \bibinfo{person}{Jiashi Feng}.} \bibinfo{year}{2021}\natexlab{}.
\newblock \showarticletitle{Deep long-tailed learning: A survey}.
\newblock \bibinfo{journal}{\emph{arXiv preprint arXiv:2110.04596}}
  (\bibinfo{year}{2021}).
\newblock


\end{thebibliography}

\end{document}